\documentclass{article}
\usepackage{spconf,amsmath,graphicx}
\usepackage{multirow}
\usepackage{amsfonts}
\usepackage{booktabs}

\usepackage{tikz}
\usepackage{etoolbox}
\usepackage{booktabs}
\usepackage{tikzsymbols}

\usepackage{algorithm}
\usepackage{algorithmic}

\newcommand{\newcite}[2]{#1 et al.~\cite{#2}}

\title{Integrating Pretrained Language Model \\ for Dialogue Policy Evaluation}
\name{Hongru Wang, Huimin Wang, Zezhong Wang, Kam-Fai Wong\sthanks{Corresponding Author} }
\address{ Department of Systems Engineering and Engineering Management\\
The Chinese University of Hong Kong }
\begin{document}
%\ninept
%
\maketitle
%
% adversarial loop paper
% huiming

\begin{abstract}

Reinforcement Learning (RL) has been witnessed its potential for training a dialogue policy agent towards maximizing the accumulated rewards given from users. However, the reward can be very sparse for it is usually only provided at the end of a dialog session, which causes unaffordable interaction requirements for an acceptable dialog agent. Distinguished from many efforts dedicated to optimizing the policy and recovering the reward alternatively which suffers from easily getting stuck in local optima and model collapse, we decompose the adversarial training into two steps: 1) we integrate a pre-trained language model as a discriminator to judge whether the current system action is good enough for the last user action (i.e., \textit{next action prediction}); 2) the discriminator gives and extra local dense reward to guide the agent's exploration. The experimental result demonstrates that our method significantly improves the complete rate (~4.4\%) and success rate (~8.0\%) of the dialogue system.

%Reinforcement Learning has been witness its potential for dialogue policy learning by modelling the user and system as environment and agent, aiming to handle complex user goals and different constrains in dialogue. However, the typical reward model in dialogue policy learning is sparse and delayed, which causes sample inefficient. Previous work mostly recovers the reward model through adversarial inverse reinforcement learning, which easily get stuck in local optima and model collapse. Instead of optimizing the policy and reward model alternatively, we decompose the adversarial training into two steps. First, we integrate pre-trained language model as a discriminator, which can predict whether or not current system action is suitable for the last user action (i.e., \textit{next action prediction}). Meanwhile, the discriminator gives and extra local dense reward to guide the agent's exploration. The experimental result demonstrates that our method significantly improves the complete rate (~4.4\%) and success rate (~8.0\%) of dialogue system.
\end{abstract}
\section{Introduction}
\label{sec:intro}
Reinforcement learning has revolutionized the way to model the dialogue policy which decides the next action of the dialogue system suited to the current state \cite{gao2018neural,walker2000application,li2009reinforcement,bacon_option-critic_2016,peng_composite_2017,takanobu_guided_2019,li_guided_2020}. On the flip side, one notorious limitation of reinforcement learning is \textit{reward sparsity} issue where here the system usually receive a positive (or negative) reward signal when the dialogue ends successfully (or unsuccessfully). Thus, the reward signal is delayed and sparse, making it extremely difficult to connect a long series of actions to a distant future reward especially for complex goals across multiple domains \cite{takanobu_guided_2019}.

% sparse reward
A typical reward function, for example, often apply a minor negative penalty (i.e., -1) in the middle of the session to encourage the system to accomplish the task in fewer turns, with a huge positive (i.e., +40) or negative (i.e., -40) reward at the end \cite{gao2018neural}. This kind of \textit{global reward} based on $(state, action)$ pairs is not informative and lead to exploration in large action space inefficient \cite{wang-etal-2020-learning-efficient}. To get more dense and enlightening reward signals, most previous works recovers the intrinsic \textit{local reward} from expert demonstrations through \textit{reward shaping} \cite{peng_composite_2017,peng-etal-2018-deep}. More specifically, some works train a discriminator to differentiate $(state, action)$ generated by dialogue agents from $(state, action)$ by expert and then regards the discriminator as a reward estimator to provide intrinsic reward signals, where the dialogue policy model and discriminator updates alternately on the fly \cite{peng2018adversarial,takanobu_guided_2019}. A further line of work decomposes the whole training into two steps by firstly training the discriminator with an auxiliary dialogue generator and secondly incorporating it into common RL method, since the alternative update mechanism limits the policy model to policy-gradient-based algorithms \cite{li_guided_2020}. However, the vast bulk of annotated $(state, action)$ pairs from expert demonstration is hard to acquire. Moreover, the reward model based on state-action pair might cause unstable policy learning and affect optimization speed with the limited amount of annotated dialogues \cite{yang2018unsupervised}.
% state action have another disadvantages which illustrate by one paper, find it. Semi-Supervised Dialogue Policy Learning via Stochastic Reward Estimation

Our work keeps in line with the methods to decompose the whole training into two sequential steps. We incorporate the pretrained language model as the reward model into common RL method to provide dense \textit{local reward} signals, guiding the action decision of dialogue policy learning. Specifically, we re-formulate one of the pre-training sub-tasks of BERT \textit{Next Sentence Prediction} as \textit{Next Action Prediction} at the first step. Given current user action $a_u$, the classifier will distinguish whether or not the response system action $a_s$ is suitable or acceptable. Intuitively, if the system chooses the right action to answer user's query at each turn, then the dialogue naturally succeeds at the end. Secondly, the trained classifier as the dialogue reward model will be incorporated into the RL process to guide the dialogue policy learning without updates. There are several advantages of our proposed method: 1) Our method is model-agnostic which can be incorporated in any RL algorithm to guide the policy learning, 2) Only action-pairs demonstration reduce the cost of annotation compared with state-action pairs demonstration, and 3) Pre-trained language model has been proved powerful and transferable in many NLP tasks which can capture the subtle difference of action-pairs, providing more dense reward signal at each turn.

% our solutions / contributions
The main contribution of this paper is two-fold: 1) we propose a simple yet effective reward estimator at the action-level to guide the action decision, and 2)
We investigate the effects of \textit{global reward} and \textit{local reward}, and the experimental results on MultiWOZ \cite{budzianowski2018multiwoz} show that our methods outperform single \textit{global reward} about 4.4\%, and furthermore, almost 8\% when combined with \textit{global reward}, which indicates the complementary effects of these two types of reward.

% global reward and local reward

% bert supervised learning

% evaluation of action (dialogue policy learning)

\section{Related Work}
\label{sec:format}
% two categories: decompose / without loop
% gdpl
% Reward shaping is a set of approaches to leverage prior knowledge to provides the agent with an extra intermediate reward $F$ (i.e., local reward) in addition to environmental reward $R$ (i.e., global reward), making the system learn from a composite signal $R + F$ \cite{ng1999policy}. The early work that provides reward shaping information in addition to the primary environmental feedback in task-oriented dialogue system is proposed by \cite{su-etal-2015-reward}, which examines three recurrent neural network (RNN) approaches at the turn-level. There has been a growing interest in exploiting adversarial training for reward shaping in dialogue policy learning \cite{DBLP:conf/icassp/PengLGLCW18,takanobu_guided_2019,wang-etal-2020-learning-efficient,li_guided_2020}.

The first focus of reward shaping is recovering the intrinsic local reward from expert demonstration by inverse reinforcement learning or adversarial training.  \newcite{Peng}{peng2018adversarial} proposed an adversarial advantage actor-critic (Adversarial A2C) method based on $(state, action)$ pairs from simulation and expert demonstration. Similarly, \newcite{Takanobu}{takanobu_guided_2019} utilizes Adversarial Inverse Reinforcement Learning to jointly estimate reward and optimize dialogue policy in multi-domain task-oriented dialog. \newcite{Wang}{wang-etal-2020-learning-efficient} directly estimate potential-based reward function from demonstrations. Nevertheless, methods alternately learning the reward model and policy cannot be extended to off-policy methods which benefit from self-learned reward functions \cite{li_guided_2020}.

The second focus of reward shaping is to train reward model and dialogue policy consecutively, rather than alternatively as mentioned above \cite{li_guided_2020}. \newcite{Li}{li_guided_2020} appropriate the reward model with multilayer perceptron (MLP) to train a discriminator based on $(state, action)$ pairs. Besides that, \newcite{Gabriel}{gordon2020learning} propose a No Label Expert (NLE) that uses an unannotated dialog dataset consisting of pairs of sentences $(s_u, s_s)$, representing user utterances and the corresponding agent responses, guiding the dialogue policy. Distinguishing from these works, we tackle \textit{reward sparsity} by incorporating pretrained language model as a reward estimator and train a discriminator at action-level rather than sentence-level or $(state, action)$ pairs.

% weak demo

% But none of this work distinguishes …, which is modeled and exploited in our work. (action-level

\begin{figure}[t]
	\centering
	\vspace{-0.3cm}
 	\footnotesize
 	\begin{tikzpicture}
 	\draw (0,0) node[inner sep=0] {\includegraphics[width=1.0 \columnwidth]{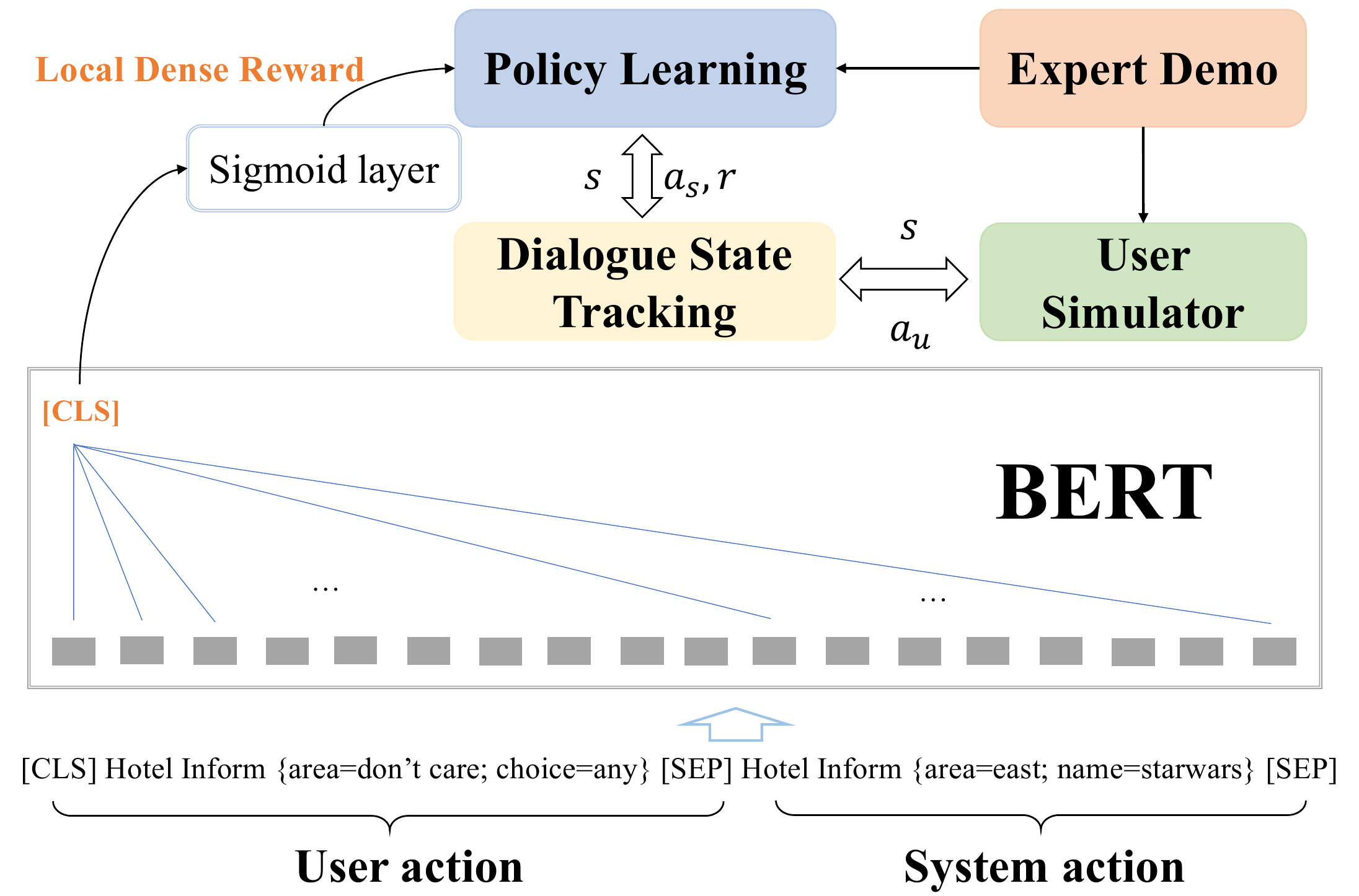}};
 	\end{tikzpicture}
 	\caption{Our proposed method: Integrating pre-trained language model into reinforcement learning as reward model to provide local dense reward signal based on action pairs $(a_u, a_s)$, guiding the optimization of dialogue policy learning.}\vspace*{-4mm}
 	\label{fig:frameowork}
\end{figure}

\section{Methodology}

In this section, we firstly introduce the detail of next action prediction (namely, how to train the discriminator), and then presents an algorithm that incorporates the reward signal given in the first step (namely, how to update the dialogue policy). Figure~\ref{fig:frameowork} shows the overall framework of our proposed method. 

\subsection{Next Action Prediction}

\textit{Next Sentence Prediction} is one of pre-trained task in BERT, aiming to predict whether, for a given pair of sentences $(s_1, s_2)$, $s_2$ is an next sentence to $s_1$. Similarly, \textit{Next Action Prediction} task is to determine whether or not the current system action $a_s$ is an appropriate response to the last user action $a_u$ in dialogue. An action (i.e., $a_u$ and $a_s$) consists of domain, intent, slots and its corresponding value as follows \cite{peng2020few}:

\vspace{-8mm}
\begin{equation}
    \mathcal{A} = [ \underbrace{~~\mathrm{D} {~_{~}}}_{\text{Domain}} \underbrace{~~\mathrm{I} {~_{~}}}_{\text{Intent}} \underbrace{ \{ \, s_1 \, = \, v_1 \, ; \cdots; \, s_P \, = \, v_P \, \} }_{\text{Slot-value pairs} } ]
\end{equation}

At each turn of multi-domain dialogue, the user and system may inquire and provide information across different domain. In this case, we represent the action as:

\vspace{-6mm}
\begin{equation}
    a_u = [ \mathcal{A}_u^1, \mathcal{A}_u^2, ..., \mathcal{A}_u^n ] \quad
    a_s = [ \mathcal{A}_s^1, \mathcal{A}_s^2, ..., \mathcal{A}_s^n ]
\end{equation}

Then, we concatenate the user action $a_u$ and system action $a_s$ and feed it into BERT. Two special tokens $[CLS]$ and $[SEP]$ are added to indicate the start and separation of actions respectively. The input representations consist of token embedding, segment embedding and positional embedding.

\vspace{-2mm}
\begin{equation}
    H = \textrm{BERT} ( \textrm{emb} (a_u, a_s))
\end{equation}

The embedding of first token (i.e. [CLS]) then is then feed into a sigmoid layer to do classification as defined:

\vspace{-4mm}
\begin{equation}
    f(x) = \mathrm{sigmoid} (Wh_1 + b)
\end{equation}

The final output of the binary classifier $f(x)$ is a probability that indicates the confident score that the system action is suitable and appreciate given last user action. The objective of the classifier $D_\phi$ can be represented as follows:

\vspace{-6mm}
\begin{equation}
    L_{D} = \mathrm{E} (log (1 - D_\phi(a_u, a_s)_{sim})) - \mathrm{E} ( D_\phi(a_u, a_s)_{real}) 
\end{equation}

After the discriminator is trained, we will keep it as the reward function for future dialogue policy learning without updates, which is illustrated at later section.

\subsection{PPO-OFF}

A trajectory $(s_0, a_0, s_1, a_1, . . . )$ is generated by sampling actions according to the policy $a_t \sim \pi(a_t| s_t)$ consecutively, until the terminal states is reached. Here, the action $a_t$ can be further divided into the user action $a_u$ and system action $a_s$ in dialogue. A reward signal $r_t$ is received at each time step. 

\vspace{-4mm}
\begin{equation}
    \mathcal{T} = [(s_0, (a_u, a_s), r_0, s_1), ..., (s_n, (a_u, a_s), r_n, s_{n+1})]
\end{equation}

where $s_t, a_u, a_s, r_t, s_{t+1}$ represents the dialogue state, user action, system action, reward and next state at time step $t$ respectively. The main objective of agent is to maximize the cumulative reward $R = \sum_0^{T} \gamma^t r_t$, where $\gamma$ is a discount factor. Given the objective function, the gradient can be computed as follows:

\vspace{-2mm}
\begin{equation}
    g = \mathbb{E} [\sum_0^T \Psi_t \nabla_\theta \mathrm{log}  \pi_\theta (a_t | s_t)]
\end{equation}

where $\Psi_t$ can be estimated with different methods. We adapt generalized advantage function $A^\pi (s_t, a_t)$ here as follows \footnote{$a_t$ refers the system action $a_s$ while $a_u$ is decided by a user simulator}:

\vspace{-4mm}
\begin{equation}
    A_t^{GAE(\gamma,\lambda)} = \sum_{l=0}^{\infty} (\gamma \lambda)^l \delta_{t+l}^V
\end{equation}

where $\delta_t^V = - V(s_t) + r_t + \gamma V(S_{t+1})$. Both $\gamma$ and $\lambda$ plays key roles in the bias-variance trade-off and serve different purposes when using an approximate value function. To stabilize the policy updates, we adapt the clipped surrogate objective as follows \cite{schulman2017proximal}.

\vspace{-4mm}
\begin{equation}
    L^{CLIP} = \mathbb{E} [min(r_t(\theta) A_t^{GAE}, clip(r_t(\theta), 1-\sigma, 1+\sigma)A_t^{GAE} )]
\end{equation}

where $r_t(\theta) = \frac{\pi_{\theta} (a_t | s_t)}{\pi_{\theta_{old}} (a_t | s_t)}$. In addition to policy, we also parametrize the value function and add entropy bonus to ensure sufficient exploration\cite{schulman2017proximal}. The whole training objective is defined as follows:

\vspace{-4mm}
\begin{equation}
     L^{CLIP+VF+S} = \mathbb{E}[ L^{CLIP} - c_1 * L^{VF} + c_2 * S[\pi](s) ]
\end{equation}

where $c_1, c_2$ are coefficients, and $S$ denotes an entropy bonus, and $L^{VF}$ is a squared-error loss $(V_\theta(s_t) - V_t^{tar})^2$.

It is obvious that the reward model plays a key role in dialogue policy learning since it directly affect the training objective. To investigate the effects of different reward signals, we replace the $r_t$ in the trajectory with three different types of reward: \textit{global reward}, \textit{local reward} and \textit{combination}.

% There are three different reward types: 1) \textit{global reward} (i.e. traditional human-defined reward), 2) \textit{local reward} (i.e. probability output from discriminator) and (3) \textit{both}. Intuitively, the combination of global and local reward signal provides more dense and informative signal compared with only local or only global, which also is verified in later experiments \footnote{It is noted that this kind of local reward can be regarded as an evaluation metric to evaluate the qualify of current system action.}.

For \textit{global reward}, we simply assign it a large positive (v.s. negative) reward +40 (v.s. -40) when the dialogue ends successfully (v.s. unsuccessfully), while -1 during the middle of session to encourage shorter session. For \textit{local reward}, we remap the output confident score to a range of [-1, 1] as follows, aiming to encourage the policy to decide more high qualified actions.

\vspace{-4mm}
\begin{equation}
    r_{local} = - 1 + 2 * D_\phi (a_u, a_s) \quad
    r_{comb} = r_{global} + r_{local}
\label{reward:adj}
\end{equation}

% \begin{equation*}
%  r_{global} =\begin{cases}
%           +c \quad &\text{if dialogue ends successfully} \,  \\
%           -c \quad &\text{if dialogue ends unsuccessfully } \, \\
%           -1  \quad &\text{if dialogue does not end}
%      \end{cases}
% \end{equation*}

\section{Experiments and Analysis}

\subsection{Dataset and Evaluation Metric}
% multiwoz
All models are evaluated on MultiWOZ \cite{budzianowski2018multiwoz}, a multi-domain, multi-intent task-oriented dialog corpus that contains 7 domains, 13 intents, 25 different slots and 10483 dialog sessions. We report the average number of \textit{dialog turns}, averaging over successful dialog sessions and all dialog sessions respectively, to measure the efficient of accomplishing a task. A dialog turn consists of a user utterance and a subsequent system (i.e. utterance pairs). \textit{Precision}, \textit{recall} and \textit{F1} are calculated based on dialog acts (i.e. action pairs) \cite{stolcke2000dialogue}. \textit{Match rate} assesses whether the offered entity meets all the constraints specified in a user goal. The dialog is marked as successful if and only if both inform recall and match rate are 1.
%For dataset, we choose MultiWOZ \cite{budzianowski2018multiwoz}, which is a multi-domain, multi-intent task-oriented dialog corpus that contains 7 domains, 13 intents, 25 different slots and 10483 dialog sessions. For evaluation metrics, we report the number of \textit{dialog turns}, averaging over successful dialog sessions and all dialog sessions respectively, to measure the efficient of accomplishing a task. A dialog turn consists of a user utterance and a subsequent system (i.e. utterance pairs). \textit{Precision}, \textit{recall} and \textit{F1} are calculated based on dialog acts (i.e. action pairs) \cite{stolcke2000dialogue}. \textit{Match rate} assesses whether the offered entity meets all the constraints specified in a user goal. The dialog is marked as successful if and only if both inform recall and match rate are 1.

\subsection{Baselines}

\textbf{MLE}: One of representative work of supervised learning to learn dialogue policy, which employs a multi-class classification via imitation learning (i.e., behavioral cloning) with a set of compositional actions where a compositional action consists of a set of dialog act items.

\noindent \textbf{GDPL}: \cite{takanobu_guided_2019} A method which alternatively updates dialogue policy and reward estimation model by using adversarial inverse reinforcement learning. It is noted that reward estimator recovers the reward signal form the state-action pairs at each dialogue turn.

\noindent \textbf{PPO} \cite{schulman2017proximal} A policy-based reinforcement learning method which uses multiple epochs of stochastic gradient ascent and a constant clipping mechanism the soft constraint to perform each policy update. The dialogue policy model in GDPL is also PPO.

\noindent \textbf{PPO-OFF-Comb} The reward model offers the combination of \textit{global reward} and \textit{local reward} while the policy updated as illustrated in section 3.2. Similarly, \textbf{PPO-OFF-Local} only receives local dense reward from pre-trained language model (i.e., BERT) and \textbf{PPO-OFF-Global} receives only global reward respectively.

\subsection{Implementation Details}

For \textit{next action prediction}, we firstly build the binary classification dataset from MultiWoZ \cite{budzianowski2018multiwoz} automatically and get 99370, 13157, 13073 labeled samples for training, testing and validation respectively \footnote{We sample user action $a_u$ with system action $a_s$ from same dialogue as positive sample, and randomly sample another system action from other dialogue to form negative sample.}. And then we use BERT\cite{devlin2018bert} as backbone and $Adam$ as optimization algorithm. The specific hyper-parameters are deployed as follows: batch size as 4, learning rate as 5e-5, epochs as 10, adam epsilon as 1e-8, the max sequence length as 512 \footnote{It is noted that the trained BERT model achieved 97.34\% accuracy at the binary classification task, which proves the pre-trained language model is capable to provide reliable reward signal.}.

\noindent For \textit{dialogue policy learning}, we adapt ConvLab-2 \cite{zhu2020convlab} as our environment and evaluate the policy at sentence-level instead of action-level. In this case, other components in task-oriented dialogue system are \textit{BERTNLU}, \textit{RuleDST} and \textit{Template-based NLG}. More specifically, we set the maximum turn of one conversation as 20, which means the dialogue will be terminated and regarded as failure when the turn excesses 20. We sample 1024 trajactories each epoch and the max epoch is set as 200. Other hyper-parameter settings follows \newcite{Takanobu}{takanobu_guided_2019}.

\begin{table}[t]
    \centering
    \small
    \begin{tabular}{l|ccc}
    \toprule[1pt]
    \multirow{2}{*}{Model} &  \multicolumn{3}{c}{Agenda} \\\cline{2-4}
      & Precision/Recall/F1 & Match & Success \\
    \hline
    \textbf{Human}  & 81.9/93.4/85.3 & 92.1 & 82.7 \\
    \hline
    MLE  & 63.1/72.6/64.5 & 50.1 & 47.0 \\
    GDPL  & 63.1/73.0/64.6 & 50.0 & 47.2 \\
    PPO & 64.4/78.3/68.2 & 64.7 & 61.2 \\
    \hline
    PPO-OFF-Global  & 63.0/81.2/67.8 & 70.8 & 63.2 \\
    PPO-OFF-Local & 60.9/82.9/67.4 & 70.9 & 67.6 \\
    PPO-OFF-Comb & \textbf{67.2}/\textbf{85.7}/\textbf{72.7} & \textbf{79.6} & \textbf{71.4} \\
    \bottomrule[1pt]
    \end{tabular}
    \caption{Performance of different dialog agents on the multi-domain dialog corpus by interacting with the agenda-based user simulator. Success average turns / all average turns}
    \label{tab:exp_res}
\end{table}

\subsection{Main Result}
Table~\ref{tab:exp_res} demonstrates the performance of different methods on the MultiWoZ dataset \cite{budzianowski2018multiwoz}. Consistent with intuition, the combination (i.e., PPO-OFF-Comb) of global and local reward reaches highest performance in both match rate (9\% improvement) and success rate (8\% improvement). Besides that, PPO-OFF-Local achieves comparable match rate but much higher success rate compared with PPO-OFF-Global. We attribute this to more semantic frame are correct in the dialogue since success rate is much harder to improve than match rate. Compared with previous methods such as PPO, our proposed methods demonstrate superior performance by improving almost 10\%.

\begin{figure}[h]
    \centering
    \begin{tikzpicture}
 	\draw (0,0) node[inner sep=0] {\includegraphics[width=0.8 \columnwidth, height=4cm, trim={0cm 0cm 0cm 0cm}, clip]{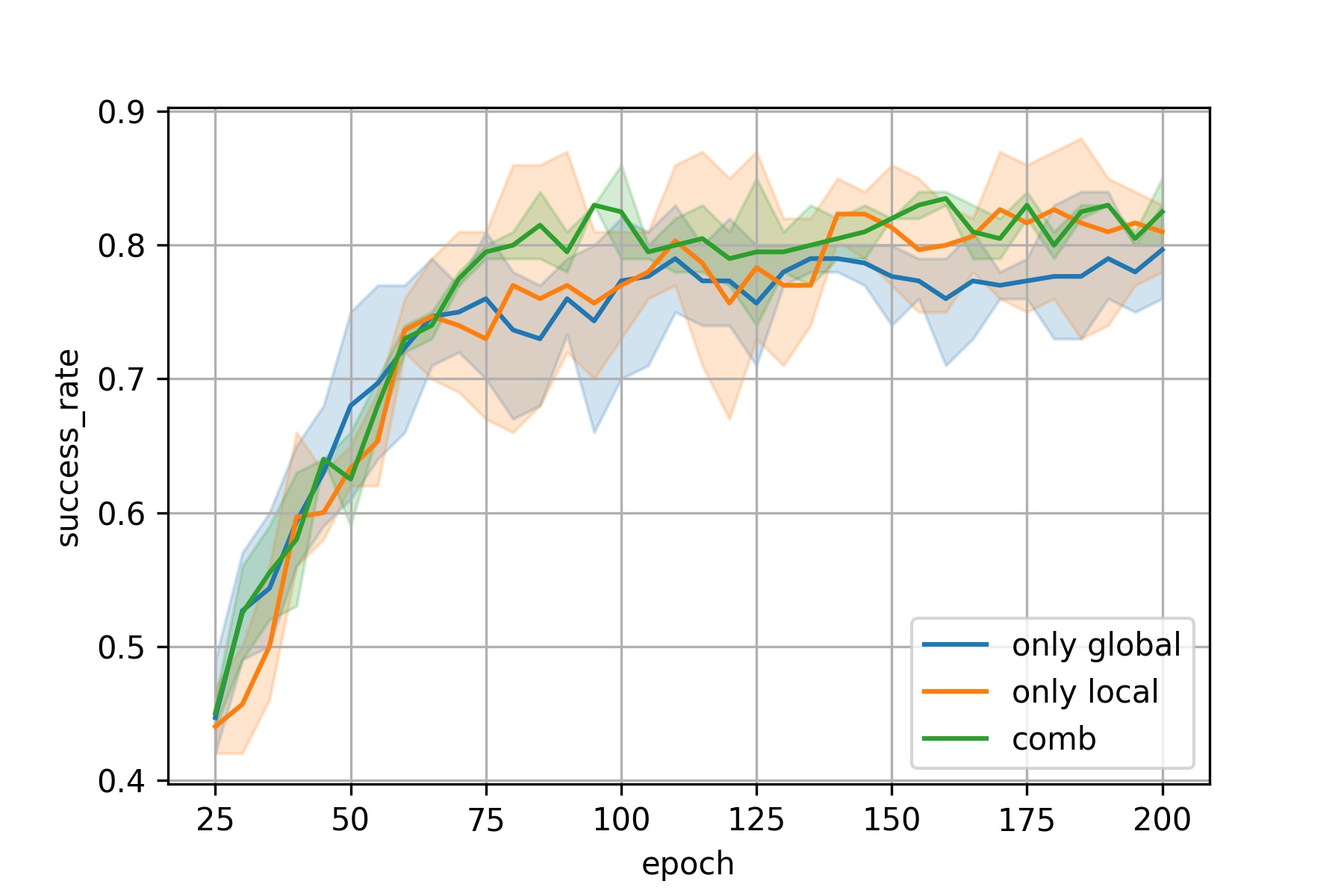}};
 	\end{tikzpicture}\vspace*{-4mm}
    \caption{Learning curves of our proposed method under different reward signals and action-level}
    \label{fig:success_rate}
\end{figure}

\subsection{Analysis}
We also conduct evaluation at the action-level (i.e., without NLU and NLG part) to learn the \textit{converge speed} and \textit{performance under different domains} respectively.

% wave, fluctuation, more stable learning
\noindent \textbf{Converge Speed} The learning curves under different reward types are presented in Fig~\ref{fig:success_rate}. We can see that the comb reward converge faster than only local and only global reward, leading to higher performance. Besides, the error band of comb reward is dramatically smaller than the other two rewards which indicates more stable policy learning. Therefore, we conclude that \textit{local reward} signals serve as a complement of \textit{global reward} to better guide and stabilize the behavior of dialogue policy. 

\begin{figure}[t]
	\centering
 	\footnotesize
 	\begin{tikzpicture}
 	\draw (0,0) node[inner sep=0] {\includegraphics[width=0.8 \columnwidth, height=3cm, trim={0cm 0cm 0cm 0cm}, clip]{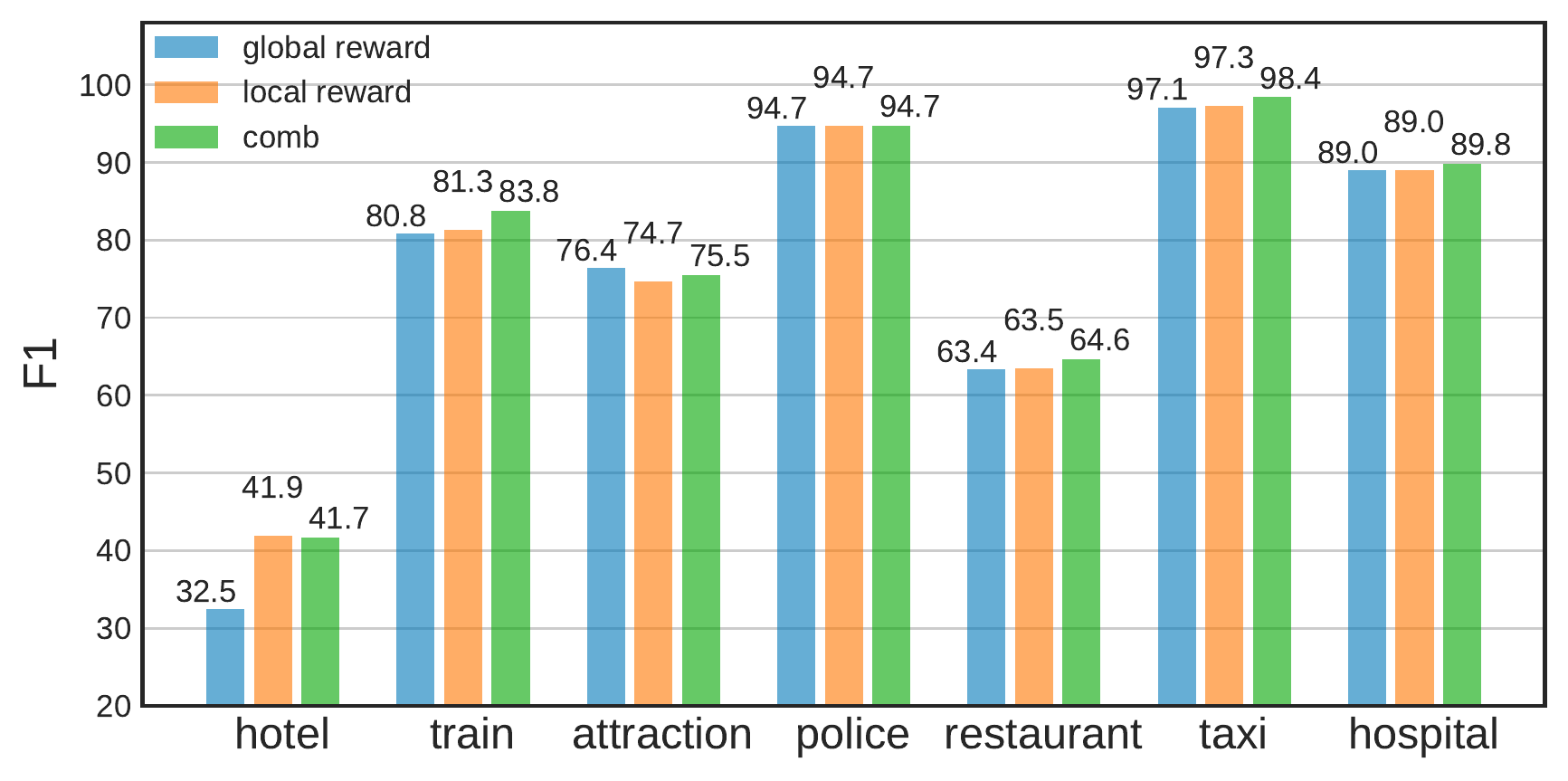}};
 	\end{tikzpicture}\vspace*{-4mm}
 	\caption{F1 score for different domain by executing policy learned with different rewards}
 	\label{fig:domain}
\end{figure}

% different domain actions
% hotel 14 restaurant 11 train 8
\noindent \textbf{The Effects of Different Domains} We also report the F1 score of different domains with policy learned from different rewards as shown in Fig~\ref{fig:domain}. The comb reward consistently outperforms single global or local reward signals, especially at the hotel, restaurant, taxi and hospital domain. We emphasize that the hotel domain contains the most slots (i.e., 14 different slots) which have the biggest improvement over such a large action space. However, we noticed that there still a large gap between different domains. We conjecture this to the data distribution of the original dataset and the difficulty and complexity of different domains.

\section{Conclusion}

In this paper, we integrates pre-trained language model (i.e. BERT) as a reward model to provide \textit{local reward} that complements with \textit{global reward}, guiding the dialogue policy learning. The experimental results demonstrate the combination of global reward and local reward reaches highest performance compares with only global or only local reward. We left more investigation such as pretraining in future work.

\section{Acknowledgements}

This work is partially supported by HK GRF\#14204118 and HK RSFS\#3133237.

\bibliographystyle{IEEEbib}
\bibliography{refs.bib}

\end{document}